\documentclass[journal]{IEEEtran}

\usepackage{hyperref}

\usepackage{array}
\newcolumntype{P}[1]{>{\centering\arraybackslash}p{#1}}

\ifCLASSINFOpdf
\else
   \usepackage[dvips]{graphicx}
\fi
\usepackage{url}

\usepackage{graphicx}
\usepackage{color}
\usepackage{xcolor}
\usepackage{amsmath}
\usepackage[flushleft]{threeparttable}

\begin{document}

\title{Non-Intrusive Detection of Adversarial Deep Learning Attacks via Observer Networks}

\author{Kirthi Shankar Sivamani, \IEEEmembership{Student Member, IEEE}, Rajeev Sahay, \IEEEmembership{Student Member, IEEE}, \\and Aly El Gamal, \IEEEmembership{Senior Member, IEEE}
\thanks{K. S. Sivamani, R. Sahay, and A. E. Gamal are with the Department of Electrical and Computer Engineering, Purdue University, West Lafayette, IN, USA. Email: \{ksivaman, sahayr, elgamala\}@purdue.edu.}}

\maketitle

\begin{abstract}
Recent studies have shown that deep learning models are vulnerable to specifically crafted adversarial inputs that are quasi-imperceptible to humans. In this letter, we propose a novel method to detect adversarial inputs, by augmenting the main classification network with multiple binary detectors (observer networks) which take inputs from the hidden layers of the original network (convolutional kernel outputs) and classify the input as clean or adversarial. During inference, the detectors are treated as a part of an ensemble network and the input is deemed adversarial if at least half of the detectors classify it as so. The proposed method addresses the trade-off between accuracy of classification on clean and adversarial samples, as the original classification network is not modified during the detection process. The use of multiple observer networks makes attacking the detection mechanism non-trivial even when the attacker is aware of the victim classifier. We achieve a 99.5\% detection accuracy on the MNIST dataset and 97.5\% on the CIFAR-10 dataset using the Fast Gradient Sign Attack in a semi-white box setup. The number of false positive detections is a mere 0.12\% in the worst case scenario.

\end{abstract}

    
\IEEEpeerreviewmaketitle

\section{Introduction}

Deep learning has achieved state-of-the-art performance in solving challenging problems such as image classification \cite{resnet}, object detection \cite{yolo}, natural language processing \cite{bert}, and automated game playing \cite{alphastar}. However, \emph{adversarial examples} \cite{intriguingProperties} have hindered the large scale deployment of deep learning models. Specifically, \emph{adversarial examples} are perturbed inputs, which are carefully crafted to induce high confidence misclassification from well-trained deep learning models. The subtle perturbations are imperceptible for a human, introducing the \emph{adversarial} aspect of these attacks. For image classification, visual imperceptibility is achieved through constrained optimization under an $L_{P}$ norm bound.

Notable properties of adversarial examples have been discovered recently that make the problem worthwhile. Most surprisingly, adversarial examples have been shown to transfer from one network to another without knowledge of the target model \cite{intriguingProperties}. Real world examples of adversarial attacks have been explored in \cite{kurakin2016adversarial} where the authors show that adversarial images retain their properties after being printed physically, or recaptured using a camera. 

Current methods in adversarial defense research follow two approaches: detection and classification. Most methods to classify adversarial examples employ deep learning techniques that can be trained end-to-end \cite{ieee_sp_4, ieee_sp_7}. This approach has two key challenges. First, the adversary may consider the defense as a part of the model, which can be attacked in the same way as the original model. Second, defenses that modify the main classification network compromise its accuracy on clean samples; this trade-off is well documented and hinders the usage of a majority of proposed defenses against adversarial attacks \cite{tradeoff}. Detection based methods have shown promising results, however, it has been shown that it is not easy to detect adversarial samples simply using a single neural network  \cite{adversarialNotEasilyDetected}. Thus, in our work, we aim to accurately detect attacks by employing multiple observer networks that classify the outputs from the hidden layers of the main network as clean or adversarial. The use of more than one observer network makes it non-trivial to attack the classification model, while keeping the original network parameters unchanged. This ensures accurate adversarial detection without compromising accuracy on clean samples. Detecting adversarial samples can be critical in systems such as self-driving vehicles, where human intervention can be sought.

Adversarial perturbations are imperceptible at the input level. However, Xie et al. \cite{Xie_2019_CVPR} show that these perturbations grow when propagated through a deep network and appear as significant noise at the hidden layers' feature maps. Motivated by this fact, we augment the main network with multiple binary classifiers to detect this amplified noise in the feature maps. Each detector takes inputs from a different hidden layer from the original classification network and classifies the input as clean or adversarial. \textbf{The parameters of the original model remain unaffected, thus retaining accuracy on clean samples}. The ensemble of all detectors is used to determine whether the input is clean or not. Most of the binary classifiers rely on convolutional layers as they are ideal to model the large number and dimensionality of feature maps. 
We note that the observer neural networks are trained independently and do not modify the parameters of the target classifier. This enables detectors to be trained and plugged-in to off the shelf neural networks. Thus, neural networks that are not mutable or have parameters that are not publicly available can be protected. 

{\bf Related Work:} We present related studies here on non-intrusive adversarial detection. Comparisons to these works have been made in Section \ref{sec:results}.
Feinman et al. \cite{artifacts} model the confidence of classifying adversarial samples by introducing Bayesian uncertainty estimates, which are available in neural networks that use dropout. These estimates can be captured through the variance of the output vectors obtained from different paths of the neural network when using dropout. They observe that this variance is much higher for adversarial inputs; this serves as a method of detection. Liang et al. \cite{adaptiveNoiseReduction} treat adversarial perturbations as a form of noise. They use scalar quantization and a smoothing spatial filter to denoise the inputs. They compare classification results of the original and denoised versions of the input to detect potential adversaries. In another study, Wang et al. \cite{mutationTesting} observe that classification results of adversarial inputs are much more sensitive to change in the parameters of the deep neural network. They introduce a metric to measure sensitivity and use a threshold value to filter input samples. We note that the methods discussed so far rely on capturing a pre-determined property, rather than relying on a machine learning model to detect anomalies. This often leads to sub-par detection performance due to generalization errors resulting from strong assumptions on data generation. Other signal processing techniques such as random sampling and uncertainty estimation have also led to promising results in non-intrusive adversarial input detection. \cite{ieee_sp_6} and \cite{sheikholeslami2019minimum} introduce a promising randomized approach for selective sampling from the hidden layers that aims at minimizing uncertainty in output classification. 
It is interesting to investigate in future work the combination of the key idea behind the proposed architectures and ensemble strategy with the sampling strategy proposed in \cite{sheikholeslami2019minimum}. 

The work of Metzen et al. \cite{metzen2017detecting} is most similar to our proposed approach. They train one auxiliary detector network that takes input from an internal layer of the deep neural network and classifies the input as clean or adversarial. The authors make several attempts to optimize the size and placement of the single detector in order to achieve optimal detection accuracy. Although circumventing such a detection algorithm is non-trivial, Carlini and Wagner \cite{adversarialNotEasilyDetected} show that it is possible in a defense blind case. Contrary to this work, we train multiple detectors to take input from and classify feature maps of all intermediate layers. We show that each hidden layer output is meaningful in discriminating the input samples. This is because of the diverse nature of adversarial noise present in the hidden layer outputs. Thus, using multiple networks to detect adversaries allows each network to classify results based on different types of perturbations found in the feature maps. These results are combined into an ensemble to achieve state-of-the-art detection accuracy. 

We achieve an average detection accuracy of 92.74\% and 90.53\% for two popular datasets across four popular attacks in the adversarial machine learning field, outperforming previously proposed detection algorithms on the same datasets. 




\section{Problem Setup}

\subsection{Datasets and Classifiers}

Our evaluations are conducted using two popular datasets: The MNIST dataset of handwritten digits \cite{mnist} and the CIFAR-10 dataset \cite{cifar}. MNIST consists of 60000 training images and 10000 test images with a dimension of $28\times28$, belonging to 10 classes (corresponding to the digits). The CIFAR-10 dataset consists of 50000 training images and 10000 test images with dimension $32\times32\times3$, belonging to 10 classes (dog, cat, frog, horse, deer, aeroplane, truck, ship, automobile, bird).

A popular ResNet-18 \cite{resnet} network is selected as a target classifier for both datasets. This is a deep 18 layer convolutional neural network that achieves a baseline classification accuracy of $93.02\%$ on CIFAR-10 and $99.00\%$ on MNIST.  

\subsection{Threat model}

There are three popular threat models in adversarial machine learning as described by Carlini and Wagner \cite{adversarialNotEasilyDetected}:

\begin{itemize}
    \item A zero knowledge or black box attacker who is not aware of the model architecture, model parameters, or the defense in place. 
    \item A perfect knowledge or white box attacker who is aware of the model architecture and parameters and also aware of the parameters and type of defense in place. 
    \item A limited knowledge or semi-white box attacker who is aware of either the model or defense.
\end{itemize}

We consider a realistic semi-white box threat model where the attacker has perfect knowledge of the weights of the model and the architecture, but not of the defense.   

\begin{figure*}
\centerline{\includegraphics[width=\linewidth]{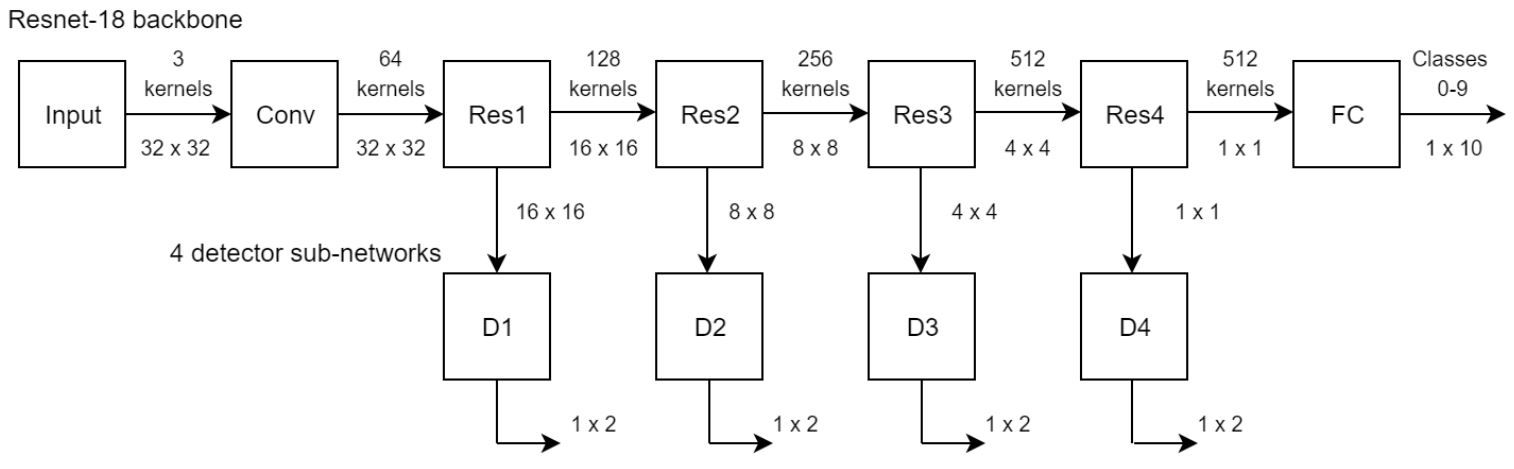}}
\caption{Defense Architecture. The boxes on the top indicate the main network (ResNet-18 \cite{resnet}) used for image classification. The \textit{Conv} block is a $7\times7$ kernel convolutional layer which is followed by 4 residual blocks (Res1, Res2, Res3, Res4). The 4 detectors (D1, D2, D3, D4) are neural networks for binary classification. They take intermediate inputs from the main network and classify the input as adversarial or clean.}
\end{figure*}

\subsection{Attacks}

We evaluate our defense on four popular attacks: Fast Gradient Sign Method (FGSM) \cite{fgsm}, Projected Gradient Descent (PGD) \cite{Madry2017TowardsDL}, DeepFool \cite{deepfool}, and the Carlini \& Wagner $L_2$ attack (CW2) \cite{Carlini_2017}. We use $x$ to denote the clean input image, $y$ for the ground truth label, $\theta$ for the network parameters, and $x^{*}$ for the constructed adversarial sample.

\textbf{FGSM} is a fast one step gradient descent with respect to the input image. The formulation of adversarial samples is:
\begin{equation} \label{fgsm_objective}
\begin{split}
    x^{*} = x + \epsilon \cdot sign(\Delta_{x}J(\theta, x, y)),
\end{split}
\end{equation}

where $\epsilon$ is the maximum allowed perturbation under an $L_{\infty}$ norm and $J$ is the loss function of the classifier. We conduct experiments using $\epsilon = 0.2$.

\textbf{PGD} can be a much stronger attack as it tolerates a higher computational cost. FGSM is performed iteratively, and the optimization step for the $k^\text{th}$ iteration is given by:
\begin{equation} \label{pgd_objective}
\begin{split}
    x^{*}_{k} = {clip_{\epsilon}}(x^{*}_{k-1} + \kappa \cdot sign(\Delta_{x}J({{\theta, x^*_{k-1}, y}}))),
\end{split}
\end{equation}

where $x^{*}_{k-1}$ is the output of the previous iteration, and $x^*_0=x$. We conduct experiments using $\kappa = 0.1$, $\epsilon = 0.2$ and 100 iterations. The `clip' function is used to clip the pixel value if the change exceeds the maximum allowed perturbation, $\epsilon$.

\textbf{DeepFool} is an iterative attack, that finds the closest hyperplane approximation of a decision boundary to $x$, and then pushes $x$ - in a perpendicular fashion - towards misclassification at the other side with minimal perturbation.




\textbf{CW2} solves the following optimization problem to find $x^{*}$:


\begin{equation}\label{cw} 
{\underset{w}{\text{min}}} \bigg|\bigg|\sigma(w) - x\bigg|\bigg|_2 ^2 + c \cdot  h\bigg(\sigma(w)\bigg), \\
\end{equation}

where $h$ is given by:

\begin{equation}
h = \text{max}(\{Z(\sigma(w))_i - Z(\sigma(w))_t, 0),\; i \neq t, 
\end{equation}

where $t$ and $Z(.)$ are the true label and softmax logit vector.

\section{Proposed Method}
We augment the main network with multiple detector observer networks. These networks take inputs from hidden layers of the main network (feature maps) and perform binary classification on these feature maps to determine whether the original input is adversarial or clean. In the considered setup, the main network is the ResNet-18 architecture that has been pre-trained\footnote{Training details: https://github.com/kuangliu/pytorch-cifar} for image classification \cite{resnet}. The parameters of the main network are frozen in order to start training the detectors. We place four detector networks that are trained independently. Similar to the main network, the detectors are also deep neural networks that end with fully connected layers down to two output classes (adversarial or clean). 
The architecture of each detector is derived from the base classification network. During training, adversarial images for the entire training set are simulated for each attack being tested. 
During inference, the four detectors are treated as an ensemble model and the input is deemed adversarial if two or more detectors classify it as so. The ensemble is meaningful because each detector learns from a different layer's feature map. This is verified by observing that the ensemble prediction of all four detectors yields a superior detection accuracy than any strict subset.  

The ResNet-18 main network has the following architecture: Input $\Rightarrow Conv \Rightarrow Res1 \Rightarrow Res2 \Rightarrow Res3 \Rightarrow Res4 \Rightarrow FC$. This is illustrated in Figure 1. Here, $ResN$ is the $N^{\text{th}}$ residual block \cite{resnet} and $FC$ is the fully connected layer that gives a $10 \times 1$ output vector. $Conv$ is a $7\times7$ stride 2 convolution with 64 kernels. Each residual block is made up of 2 $3\times3$ convolutions each followed by batch normalization and ReLU activation. The architectures of the detectors are given by: 
\begin{equation}
\begin{split}
D1: Res2 \Rightarrow Res3 \Rightarrow Res4 \Rightarrow FC \Rightarrow FC', \\
D2: Res3 \Rightarrow Res4 \Rightarrow FC \Rightarrow FC', \\
D3: Res4 \Rightarrow FC \Rightarrow FC', \\
D4: FC \Rightarrow FC'.
\end{split}
\end{equation}

Here, $FC'$ is a fully connected layer that goes from 10 to 2 neurons. The detectors are trained using the Adam Optimizer \cite{adam} with a learning rate of 0.01, batch size of 256 and a weight decay of $10^{-5}$. To the best of our knowledge, this architecture is unique and novel. The detector networks are derived from splitting the main classification network in various layers. 
Reusing the architecture in this manner leads to promising results, as shown in the next section. The bounds of the adversarial perturbations used throughout the experiments ($\epsilon = 0.2, \kappa = 0.1$) are chosen based on studies performed by similar works and results we compare to. These bounds are also low so as to make the generated adversarial examples imperceptible. Further, the bounds are kept constant throughout the training and testing time. As the adversarial examples are generated during training, the detectors are trained on the specific attack that the classifier is trying to defend against. Transferability of the proposed detection mechanism to different attacks unseen during training is left for future work.

\section{Results}\label{sec:results}

We demonstrate the effectiveness of our defense by measuring the accuracy of detection among adversarial samples (i.e., the percentage of adversarial samples that were successfully detected). We also measure the number of false positives (i.e., the number of clean images that were classified as adversarial). Across the four attacks and two datasets used, the classification accuracy post-attack (undefended network) ranged from 14\% to 22\%. Tables I and II summarize the results of adversarial detection obtained across all testing scenarios. The columns D1, D2, D3, and D4 give the accuracies of the individual binary detectors. The ensemble column gives the total detection accuracy of the defense\footnote{Code accepted for publication at:\\ \url{https://codeocean.com/capsule/3959338/tree/v1}}.

\subsection{Results on MNIST}

We achieve a detection accuracy of 99.5\% using FGSM and 89.1\% on the CW2 attack. The complete results for the attacks are shown in Table 1. False positives rates were 0.12\%, 0.08\%, 0.09\% and 0.07\% for FGSM, PGD, CW2 and DeepFool, respectively. \cite{artifacts} achieves an accuracy of 92.2\% and \cite{adaptiveNoiseReduction} achieves an accuracy of 93.86\% for detecting the FGSM attack. \cite{mutationTesting} achieves an accuracy of 97.67\% accuracy for FGSM and 94.00\% accuracy for CW2. The accuracies obtained in our experiments are superior in all but one case to these values. The reason for the lower detection accuracy compared to \cite{mutationTesting} on CW2 is left for future work. 

\begin{table}
  \caption{Detection accuracy on the MNIST dataset (\%)}
  \centering
  \begin{tabular}{|P{1.10cm}|P{1cm}|P{1cm}|P{1cm}|P{1cm}|P{1.15cm}|} 
    \hline
    Attack & D1 & D2 & D3 & D4 & Ensemble \\ \hline
    FGSM & 91.2 & 95.8 & 95.0 & 98.0 & 99.5 \\
    PGD & 82.0 & 84.0 & 84.0 & 87.4 & 91.0 \\
    CW2 & 82.5 & 86.0 & 85.8 & 86.6 & 89.1\\
    Deepfool & 83.1 & 82.5 & 86.0 & 88.5 & 91.4 \\
    \hline 

\end{tabular}
\end{table}

\begin{table}
  \caption{Detection accuracy on the CIFAR-10 dataset (\%)}
  \centering
  \begin{tabular}{|P{1.1cm}|P{1cm}|P{1cm}|P{1cm}|P{1cm}|P{1.15cm}|} 
    \hline
    Attack & D1 & D2 & D3 & D4 & Ensemble \\ \hline
    FGSM & 89.0 & 94.4 & 94.0 & 95.5 & 97.5 \\
    PGD & 69.5 & 78.9 & 82.0 & 82.4 & 85.4 \\
    CW2 & 65.5 & 73.0 & 77.0 & 80.1 & 85.0\\
    Deepfool & 72.1 & 84.5 & 84.5 & 87.9 & 94.2 \\
    \hline 

\end{tabular}
\end{table}

\subsection{Results on CIFAR-10}

As shown in Table II, we achieve a detection accuracy of 97.5\% using FGSM and 85.0\% using CW2. False positive rates were 0.06\%, 0.02\%, 0.03\% and 0.02\% for FGSM, PGD, CW2 and DeepFool, respectively. \cite{artifacts} achieves an accuracy of 74.7\% and \cite{metzen2017detecting} achieves an accuracy of 84.00\% for detecting the FGSM attack. \cite{mutationTesting} achieves an accuracy of 91.00\% accuracy for FGSM and 83.00\% accuracy for CW2. We obtain better accuracies than these present methods of detection. 

\subsection{False Positives}

Many present methods of adversarial defenses are rendered unusable due to their drop in accuracy on clean samples \cite{tradeoff}. We address this trade-off by keeping the original classification network unchanged. For all testing scenarios, we observe the number of clean testing samples in MNIST and CIFAR-10 that were classified as adversarial. In the worst case scenario, there were a total of 12 clean images classified as adversarial (false positive rate of 0.12\%). These results validate our hypothesis of minimal loss in accuracy on clean samples.

\subsection{Ablation Study}

To assess the impact of each detector on the final detection accuracy, several experiments were performed. Tables I and II give the accuracy of detection for each individual detector as well as the ensemble when using all four detectors for MNIST and CIFAR-10. The ensemble accuracy is consistently higher than that of any individual detector and this shows the effectiveness of the ensemble result.  
Experiments were also conducted to verify the need for multiple detectors. Ensemble accuracies were calculated using two subsets of the four detectors: D1 + D4 (peripheral: closer to input and output) and D2 + D3 (middle: farther from the input/output). In these cases of two detectors, the input is considered adversarial when both detectors classify it as so. Table III shows the accuracy of detection under these scenarios. The inferior accuracy compared to the ensemble of all four detectors demonstrates the need for each detector. This strongly suggests that each detector learns new discriminative features about the adversarial image from the output of various hidden layers. Results using other combinations of D1 through D4 are provided in the Appendix.

\begin{table}
  \caption{Ensembling combinations (\%)}
  \centering
  \begin{tabular}{|P{1.32cm}|P{1.32cm}|P{1.32cm}|P{1.32cm}|P{1.32cm}|} 
    \hline
    Attack & Dataset & D1 + D4 & D2 + D3 & Ensemble \\ \hline
    FGSM & MNIST & 98.5 & 97.5 & 99.5 \\
    PGD & MNIST & 88.5 & 90.4 & 91.0 \\
    CW2 & MNIST & 87.1 & 87.5 & 89.1 \\
    Deepfool & MNIST & 87.8 & 90.0 & 91.4 \\
    FGSM & CIFAR-10 & 96.5 & 96.0 & 97.5 \\
    PGD & CIFAR-10 & 81.4 & 83.5 & 85.4 \\
    CW2 & CIFAR-10 & 81.3 & 82.5 & 85.0 \\
    Deepfool & CIFAR-10 & 88.5 & 92.2 & 94.2 \\
    \hline 

\end{tabular}
      \begin{tablenotes}
      \item This table shows the accuracy of detection for 2 different combinations of ensembling the detectors. Combination D1 + D4 is of the peripheral detectors and the combination D2 + D3 is of the middle detectors. The number of false positives (FP) is given over testing on 10000 images. Here, the false positives are calculated over ensembling all 4 detectors.
    \end{tablenotes}
\end{table}

\section{Discussion and Future Work}

Our defense does not incorporate any baseline defense such as adversarial training~\cite{fgsm}. Adversarial training is a widely used defense, which trains the classifier using adversarial samples and their corresponding corrected labels. Detection methods commonly employ this technique and build a defense on top of it. However, this is not useful for our detection based method, because we seek to retain discriminative properties of adversarial and clean samples. This means that adversarial training modifies the parameters of the network to improve classification of adversarial samples, which in turn reduces the differences between outputs of hidden layers for adversarial and clean inputs and thus, making them harder to detect. 

The observer networks that we employ in our method of defense have two main properties that make it effective: non-intrusiveness and diversity. The non-intrusive aspect of the detection is that the observer networks employed are trained independently from the original classifier. Thus, the weights and gradients of the observer networks differ from those of the main network. The non-intrusive nature of our proposed detection mechanism allows for simple deployment on existing models. The diversity aspect of detection is that we use multiple observer networks for detection of an adversary. This allows us to take inputs from various layers of the neural network and classify them. These inputs (feature maps) are different from each other (diverse) and enable different bases of classifying the input as adversarial or clean. 


We achieve state-of-the-art detection results on the MNIST and CIFAR-10 datasets against powerful iterative attacks such as PGD and CW2. However, application of this defense on datasets of higher resolution using larger network architectures is open for further study. 
In particular, we anticipate exploring the scalability of our defense to deeper models. 


\newpage

\bibliography{ref}
\bibliographystyle{IEEEtran}

\appendix
\section*{Other Detector Combinations}
Table I shows the accuracies of using an ensemble of two other combinations of detectors D1 through D4 than the ones mentioned in the main text. In the case of D2 to D4, three detectors D2 + D3 + D4 are used for the ensemble and the majority classification result is selected (clean or adversarial).

\begin{table}
  \caption{Ensembling combinations (\%)}
  \centering
  \begin{tabular}{|P{1.32cm}|P{1.32cm}|P{1.32cm}|P{1.32cm}|P{1.32cm}|} 
    \hline
    Attack & Dataset & D3 + D4 & D2 to D4 & Ensemble \\ \hline
    FGSM & MNIST & 99.0 & 97.9 & 99.5 \\
    PGD & MNIST & 88.1 & 89.5 & 91.0 \\
    CW2 & MNIST & 87.1 & 84.5 & 89.1 \\
    Deepfool & MNIST & 87.0 & 91.0 & 91.4 \\
    FGSM & CIFAR-10 & 96.0 & 95.2 & 97.5 \\
    PGD & CIFAR-10 & 80.5 & 84.0 & 85.4 \\
    CW2 & CIFAR-10 & 79.9 & 83.0 & 85.0 \\
    Deepfool & CIFAR-10 & 88.6 & 92.1 & 94.2 \\
    \hline 

\end{tabular}
      \begin{tablenotes}
      \item This table shows the accuracy of detection for 2 different combinations of ensembling the detectors.
    \end{tablenotes}
\end{table}

\section*{Adversarial Noise Propagation}
Figure 1 demonstrates the propagation and amplification of adversarial perturbations through the considered ResNet-18 network. The images on the top from left to right are that of a clean and adversarial 3, randomly selected from the MNIST dataset. The adversarial image is generated using PGD with $\epsilon$ = 0.2. The images on the bottom are that of their respective feature maps obtained from the output of the Res1 layer. It can be clearly seen that the feature map of the adversarial 3 is significantly noisier than the clean 3. This motivates our approach of detection by filtering out input images with noisy feature maps in the main classification network.

\begin{figure}
\centerline{\includegraphics[width=4cm]{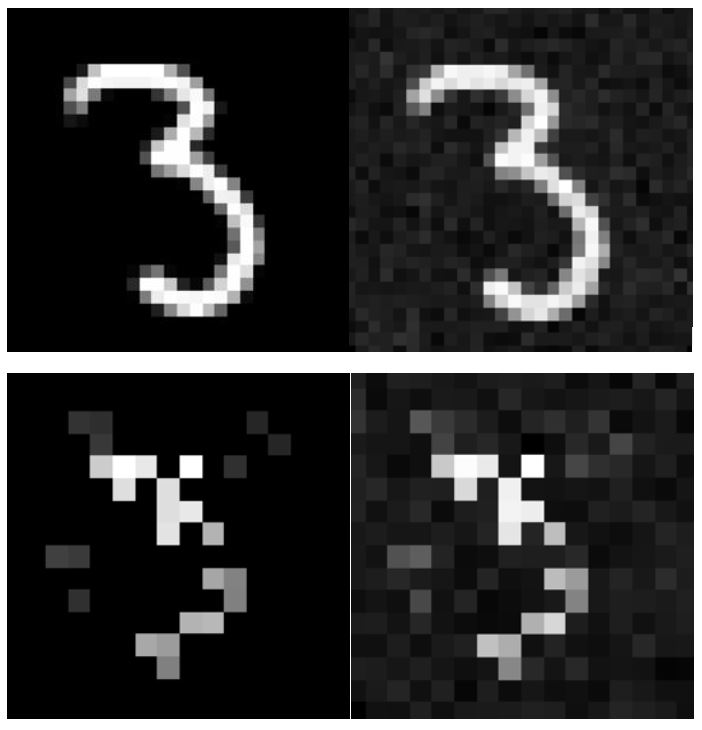}}
\caption{The images in the top row ($28\times28$) correspond to that of a clean 3 and an adversarial 3 from the MNIST dataset (left to right). The images in the bottom row ($16\times16$) are that of their corresponding feature maps after the Res1 layer of our main classification network (left to right).}
\end{figure}


\end{document}